\documentclass[letterpaper, 10 pt, journal, twoside]{IEEEtran}
\ifCLASSINFOpdf
\else
\fi

\usepackage{graphicx}
\usepackage{amsmath}
\usepackage{amsfonts}
\usepackage{bm}
\usepackage{booktabs}

\newcommand{\eqqref}{Equation~\ref}
\newcommand{\secref}{Section~\ref}
\newcommand{\figref}{Figure~\ref}
\newcommand{\tabref}{Table~\ref}

\begin{document}
%
\title{Task-oriented Motion Mapping on Robots of Various Configuration using Body Role Division}
%
%
%

\author{Kazuhiro Sasabuchi$^{1}$, Naoki Wake$^{1}$, and Katsushi Ikeuchi$^{1}$%
\thanks{Manuscript received: August, 3, 2020; Revised October, 17, 2020; Accepted November, 21, 2020.}
\thanks{This paper was recommended for publication by Editor Tamim Asfour upon evaluation of the Associate Editor and Reviewers' comments.
}
\thanks{$^{1}$All authors are with Applied Robotics Research, Microsoft, Redmond, WA, USA
        {\tt\footnotesize Kazuhiro.Sasabuchi@microsoft.com}}%
\thanks{Digital Object Identifier (DOI): see top of this page.}
}
%
%

\markboth{IEEE Robotics and Automation Letters. Preprint Version. Accepted November, 2020}
{Sasabuchi \MakeLowercase{\textit{et al.}}: Task-oriented Motion Mapping on Robots of Various Configuration using Body Role Division} 

%



\maketitle

\begin{abstract}
Many works in robot teaching either focus only on teaching task
knowledge, such as geometric constraints, or motion knowledge, such as
the motion for accomplishing a task. However, to effectively teach a
complex task sequence to a robot, it is important to take advantage of both task
and motion knowledge.
The task knowledge provides the goals of each individual task within the sequence and reduces the number of required human demonstrations,
whereas the motion knowledge contain the task-to-task constraints that would otherwise require expert knowledge to model the problem.
In this paper, we propose a body role division
approach that combines both types of knowledge using a single human demonstration.
The method is inspired by facts on human body motion and uses a body structural analogy to decompose a robot's body configuration into different roles: body parts that are dominant for imitating the human motion and body parts that are substitutional for adjusting the imitation with respect to the task knowledge. 
Our results show that our method scales to robots of different number of arm links, guides a robot's configuration to one that achieves an upcoming task, and is potentially beneficial for teaching a range of task sequences.
\end{abstract}


%
\IEEEpeerreviewmaketitle

\section{Introduction}
\label{introduction}
%
%
%
%
\IEEEPARstart{I}{n} robot teaching, one way to teach a manipulation task is to provide the geometric constraints involved in the task.
However, when a task is part of a longer multi-step sequence of tasks, this approach
lacks important information on how to complete the task in a way that takes into account the entire sequence. For example, when grasping an item from a closed cabinet, a robot must first reach the door handle position of the cabinet but do so in a way that enables achieving two following tasks: opening the cabinet, and then reaching inside with another arm. Planning a motion with the entire sequence in mind is essential but doing so with a general motion planner requires a very complex modeling of the task sequence.

Meanwhile, humans are able to move their body with the entire sequence in mind. Thus, the issue with modeling the task sequence can be relaxed if human motion mimicking is incorporated in the motion planning. The question is: how to integrate motion mimicking with the planning of the geometric task constraints.

   \begin{figure}[t]
      \centering
      \includegraphics[width=\columnwidth]{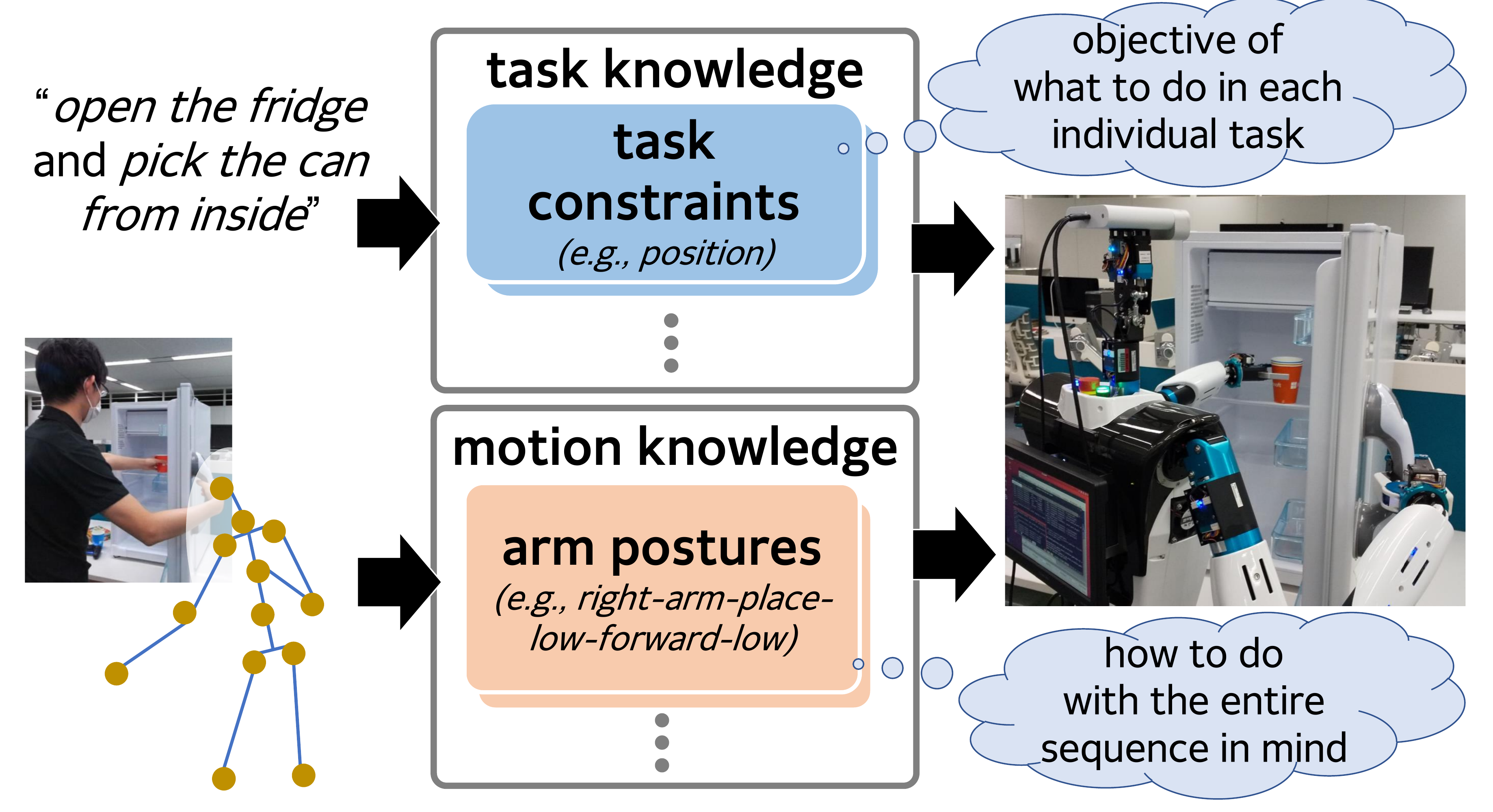}
	 \caption{An example requiring the proposed method. The constraints of each individual task (``open" and ``pick") can be defined a priori, but the task-to-task constraints (open in a way so that the other arm can reach inside for a pick) depends on the combination or the context in which the task is performed. By combining human demonstration, the task-to-task constraints are implicitly incorporated to the motion plan and thus, a feasible motion is generated to perform the multi-step sequence.}
      \label{concept}
   \end{figure}

We approach this integration problem by mimicking only the human arm motion instead of the whole-body motion, and by dividing the robot configuration into two groups: a group corresponding to the human arm which follows the human arm motion, and a remaining group that solves the task constraints. The reason we focus on the arm motion and use such grouping is that the strategy allows us to use the structural analogy of the human body motion: the human arm has control over the motion, and the human trunk acts as a range substitution \cite{kaminski1995coordination}. Since the arm is the most dominant part of the human body motion, the arm motion should provide us many valuable hints for accomplishing a task 
under a sequence of tasks.

Our main contributions in this paper are as follows. First, we present a motion planning method that integrates both task constraints of each individual task (e.g., open a door) and human motion mimicking, where the mimicking helps to incorporate the task-to-task constraints that arise when a task is performed under some context or multi-step sequence (e.g., open in a way that prepares the robot for picking from inside).
Second, we present how the method scales to robots of various configuration; including robots with fewer and equal arm links compared to the human arm. Third, we prove the effectiveness of our approach on several task sequences by looking at cases with each of the two type of robots.

%

\section{Related Works}
\label{related_works}

Robot teaching is a wide area of research.
Approaches using motion include a direct joint-to-joint mapping of the human body motion \cite{dariush2009online} or a kinesthetic demonstration \cite{argall2009survey} which teach by passively moving the robot's joints. Recent works combine kinesthetic teaching with correction \cite{bajcsy2018learning}.
In most of these approaches, observed trajectories from demonstration are mapped to mathematical models using Hidden Markov models, Gaussian mixture models, combinations of these models \cite{lee2011incremental}, and/or by fitting to dynamic or probabilistic equations \cite{schaal2006dynamic} \cite{campbell2017bayesian}.
However, kinesthetic demonstrations are difficult to scale
when the manipulation requires a mobile base movement.
Few works have tackled the problem of mapping or automatically achieving mobile base movement from demonstration. 
Welschehold et al. approached the problem by mapping the human torso movement for a single arm task \cite{welschehold2017learning}. They have further extended their work to learn individual actions and disambiguate overall task goals in a mobile setting from a small number of demonstrations \cite{welschehold2019combined}.

One limitation to these approaches is that they require repeating the demonstration several times \cite{billard2008survey}.
While obtaining task knowledge and intent completely from demonstration is a motivated goal, other works try to explicitly teach constraints so that what is taught by the human is guaranteed and mismatch in mental models \cite{akgun2012trajectories} are avoided. These
approaches focus on the symbolic or geometric constraints of the task.
Early works in this literature have focused on object state transitions \cite{ikeuchi1994toward}.
More recent works combine geometric constraints with keyframes in a multi-step task demonstration \cite{perez2017c}. Many of these approaches focus on the end-effector movements and assume that the task constraints can be solved by a general motion planner. 
However, this is only true when the robot has enough reachability, which is not the case with domestic robots with a compact structure \cite{yamamoto2019development}. An appropriate arm motion integrated with full body positioning is essential for achieving a complex task sequence with such domestic robots. In such case, additional constraints arise and must be foreseen to appropriately plan a motion under a sequence.

In this context, our motivation is to use demonstrated motions to implicitly augment these constraints. By combining the task knowledge (e.g., instructions of what to manipulate) of each individual task, we avoid requiring multiple repeating demonstrations to learn the teacher's intent; and by combining motion knowledge from a single demonstration, we avoid having to model (using expert knowledge) every ambiguity for planning a motion suitable within the task sequence.

On the line of combining task and motion knowledge, recent works target the use of language to combine task instructions with teleoperated demonstrations \cite{lynch2020language}. Due to the dependency on teleoperation, instructing and teaching complex tasks requiring multiple arms or mobile base movement is yet a challenge.
Meanwhile, works that focus on full body motion is found in computer graphics \cite{yamane2010animating}.
However, the main goals in computer graphics is to solve the constraints between the agent and the floor.
A motion is not leveraged to perform a sequence of tasks but is used to design specific character movements.


\section{Problem Definition}
\label{prerequisites}

In our problem, we assume some defined set of tasks where each task has a defined list of task constraints the robot must fulfill to perform the task. For example, let's assume we have a task set: \textit{reaching}, \textit{picking}, and \textit{door-opening}. To perform a \textit{reaching} or \textit{picking} task, the robot's end-effector must end or start at a grasp position of the grasp target; to perform a \textit{door-opening} task, the end-effector must follow the constrained movement of the door (i.e., a door can only move around the axis of the hinge).

On top of these hard constraints that must be fulfilled in each individual task, we have additional constraints that arise when the tasks are combined as a multi-step sequence. For example, to achieve a reach and pick after opening a fridge, an additional constraint to keep the robot positioned in front of the fridge is required. Instead of directly defining the additional constraints (which varies per sequence and would require careful understanding of each multi-step problem) we use a demonstration containing the human arm motion to perform the task sequence. Although not guaranteed, the additional constraints are likely to be (implicitly) contained in the arm motion as the human should have demonstrated his or her motion with the entire task sequence in mind. In the experiment section, we show several examples where the na\"{i}ve approach is yet capable of solving multi-step problems.

\figref{concept} shows an example application \cite{wake2020verbal}\cite{wake2021learning} that has our problem setting. The human shows a demonstration of the task sequence alongside with a verbal instruction of what tasks the robot should perform (a \textit{reaching} task inserted whenever specified a target object). Although the instructions provide the list of tasks to perform, the additional constraints are not fully explained and must be extracted from the demonstration.

Below we explain in more detail the task constraints and the arm motion we use for planning the robot motion.


\subsection{Task Constraints}
\label{prerequisites_task}

For the task constraint, we consider a desired position goal state (\textbf{task position goal}) $\bm{p}$ 
of the robot's end-effector, where the position goal state is the state achieved by an action $\Delta \bm{p}$ of the end-effector in a motion free direction. 
Many tasks contacting or manipulating a rigid object require solving a list of goal states
(e.g., pick, place, door opening, pressing a button, operating a kitchen faucet, etc.). 
A random solution to perform the action $\Delta \bm{p}$ may drastically change the robot's arm configuration from a mimicked motion.
Therefore, solving both a task position goal and mimicking human arm motion is an interweaved problem.

Another constraint is the orientation of the end-effector (\textbf{orientation goal}) when performing the end-effector action. The orientation goal is usually defined from the properties (e.g., the shape) of a manipulating object, but it must sometimes be obtained from the human demonstration to consider the entire task sequence. An example case is when determining to either grasp a can drink from the side or top in a picking task. Side grasping is appropriate when placing the can drink inside a shelf, whereas top grasping is appropriate when placing it inside a basket.
The type of grasp to use depends on the properties of the placing location, which is not a direct task constraint of the reaching/picking task but rather an information from the entire task sequence. We will assume that
the decision to obtain the orientation goal from the object property or human demonstration is defined a priori in a database about the manipulating object.

\subsection{Human Arm Motion}
\label{prerequisites_configuration}

\begin{figure}[t]
      \centering
      \includegraphics[width=\columnwidth]{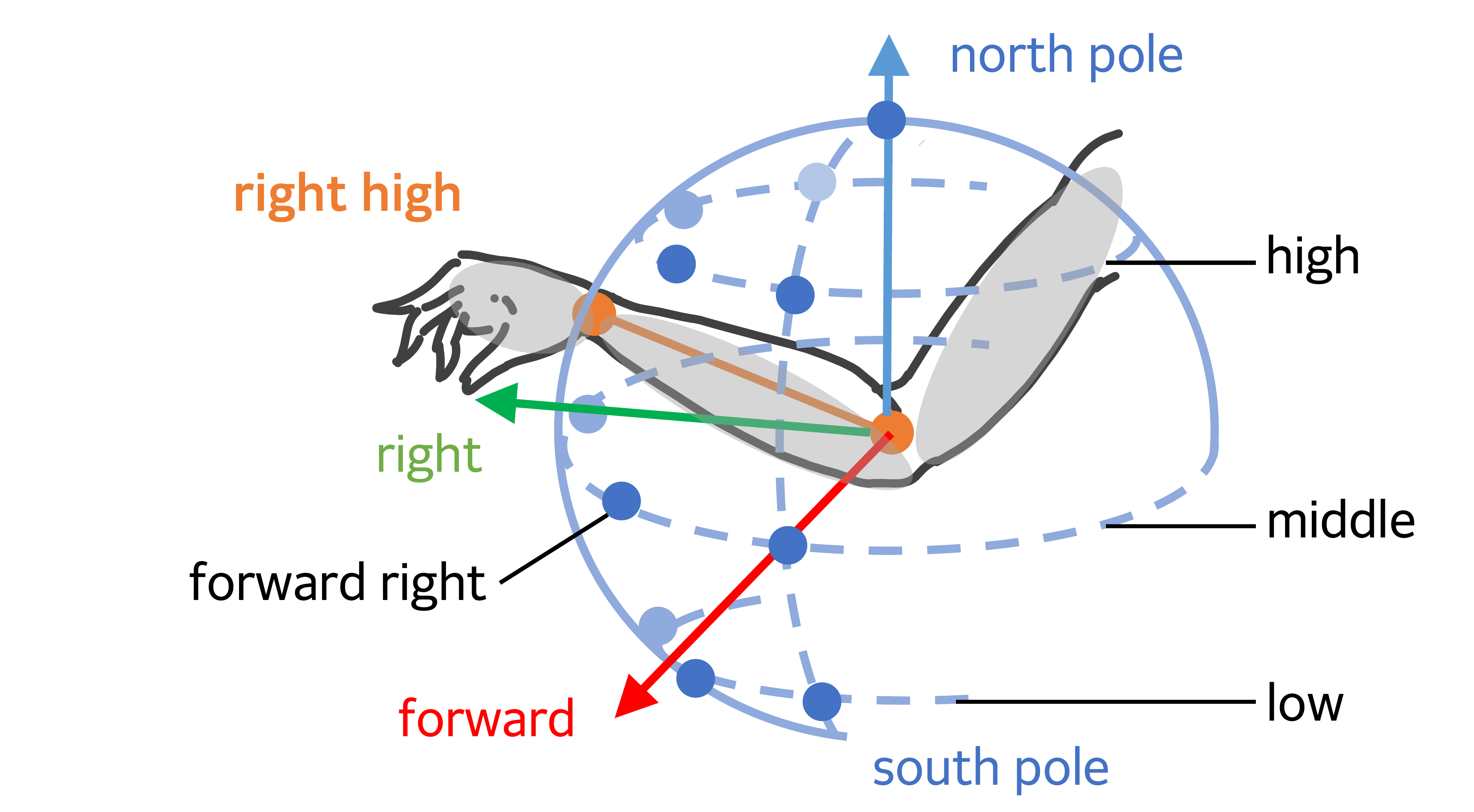}
      \caption{An eight-by-five discretized direction space expressing the pointing direction of the human upper arm and forearm. The figure shows an example of the right forearm pointing in a right high direction (forward is defined by the body facing direction at the beginning of a task).}
      \label{motion_representation}
   \end{figure}

To mimic human motion, we consider a list of desired arm postures (\textbf{arm posture goal}) represented in some intermediate representation, such as the name of the posture. To cover a variety of arm postures, we name a motion for every possible combination of the human upper arm and forearm pointing directions in some discretized direction space (\figref{motion_representation}). In this paper, we will use a direction space used in existing human motion representations \cite{guest2014labanotation}\cite{ikeuchi2018describing}, where the direction is divided into eight horizontal directions (forward, left forward, left, ...) and five vertical directions (south pole, low, middle, high, north pole). To our survey, when limited to in-front single arm manipulations, the number of valid direction combinations of the human upper arm and forearm (the number of named postures) is 79 using this eight-by-five discretization.

Another way to represent an arm posture is to represent the pointing direction of the upper arm and forearm each as a  vector of continuous float values. However, a continuous data representation will contain
noise in the raw data and will require a well-defined set of equations to convert to a robot motion. Meanwhile, an intermediate representation allows us to define a finite number of mapped configurations a priori and is able to filter noisy jumps or obvious detection errors in the human motion (e.g., unnatural arm-twisted postures can be checked a priori in a discretized representation and then be defined as unacceptable).

\section{Body Role Division Method}
\label{method}

In this section, we explain our method for planning a robot motion that satisfies both the task constraints and the mimicking of the human arm motion. By mimicking the human arm motion, we expect to achieve a robot motion with the entire task sequence in mind.

As explained in \secref{introduction}, we divide the robot configuration $\bm{q}$ into two groups: a configurational group $\bm{q^c}$ that are the joints that map the human arm motion; and a positional group $\bm{q^p}$ that are the remaining joints that solve the task constraints (desired end-effector states). An exceptional constraint that is not solved with the positional group is the orientation goal from \secref{prerequisites_task}. The orientation goal is sometimes obtained through human mimicking; thus, the group solving the goal (the orientational group) should be a subset of the configurational group.

The \textbf{configurational group} corresponds to the arm-to-hand (upper arm, forearm, and wrist-to-hand) on the human body. In this paper, we will consider robots that have an arm attached to some base, and an end-effector attached to the end of the arm. In most cases, the configurational group is the arm and the end-effector.

The \textbf{positional group} corresponds to the trunk of the human body, which is the waist or torso of the robot, but since simple-structured robots may not have such structure (or not enough joint range as the human), we will also include the robot's base. Note that, the base movement does not correspond to the human footsteps, but instead it is positioned to substitute the arm movement (\secref{mapping_less}).
To integrate mobile base movement, we will consider base movements as part of the robot joint configuration by defining a virtual prismatic and/or revolute joint attached to the robot’s base.

The \textbf{orientational group} corresponds to the wrist-to-hand part of the human arm. This is usually the wrist and end-effector on the robot, which indeed by itself is able to solve the orientation goal if had (and mostly do) three or more degrees of freedom \cite{singh2010analytical}. 
This subset of the configurational group is required due to the characteristics of the orientation goal, but 
structural facts on the human motion also insist that the human wrist motion is independent from the upper and forearm motion \cite{lacquaniti1982coordination}; therefore, defining a subset is also reasonable from a structural analogy perspective.

Using this idea of body role division, which decomposes the robot body into a configurational group, positional group, and an orientational group, we solve the arm posture goal, task position goal, and orientation goal from \secref{prerequisites}. For simplicity, we will begin with the case of a single arm posture goal, a single task position goal, and a single orientation goal (e.g., the moment of grasping). Our method uses a step-by-step calculation on each role group, which is described below:
\begin{enumerate}
\item Map the arm posture goal to a mapped configuration $\bm{{q_0}^c}$ which define a set of joint values for the configurational group. Set some predefined default configuration $\bm{{q_0}^p}$ (e.g., zero values) for the positional group.
\item By changing the joint values in the orientational group, modify $\bm{{q_0}^c}$ to joint configuration $\bm{{q_1}^c}$ which satisfies the orientational goal $\Omega_{ogoal}$.
\item Find a final configuration $\bm{q}$ which satisfies the task position goal $\Omega_{pgoal}$ by mainly changing the joint values in the positional group, but also by making sure that the configurational group is maintained using a configuration constraint $\Omega_{ccons}$, and a group connection constraint $\Omega_{pcons}$.
\end{enumerate}

The search of a configuration in the last two steps can be done by applying the goals and constraints as a fitness function in a genetic algorithm \cite{starke2017memetic}. We explain the details of each step in each of the below subsections.

\subsection{Mapping the Arm Posture Goal}

\begin{figure}[t]
      \centering
      \includegraphics[width=\columnwidth]{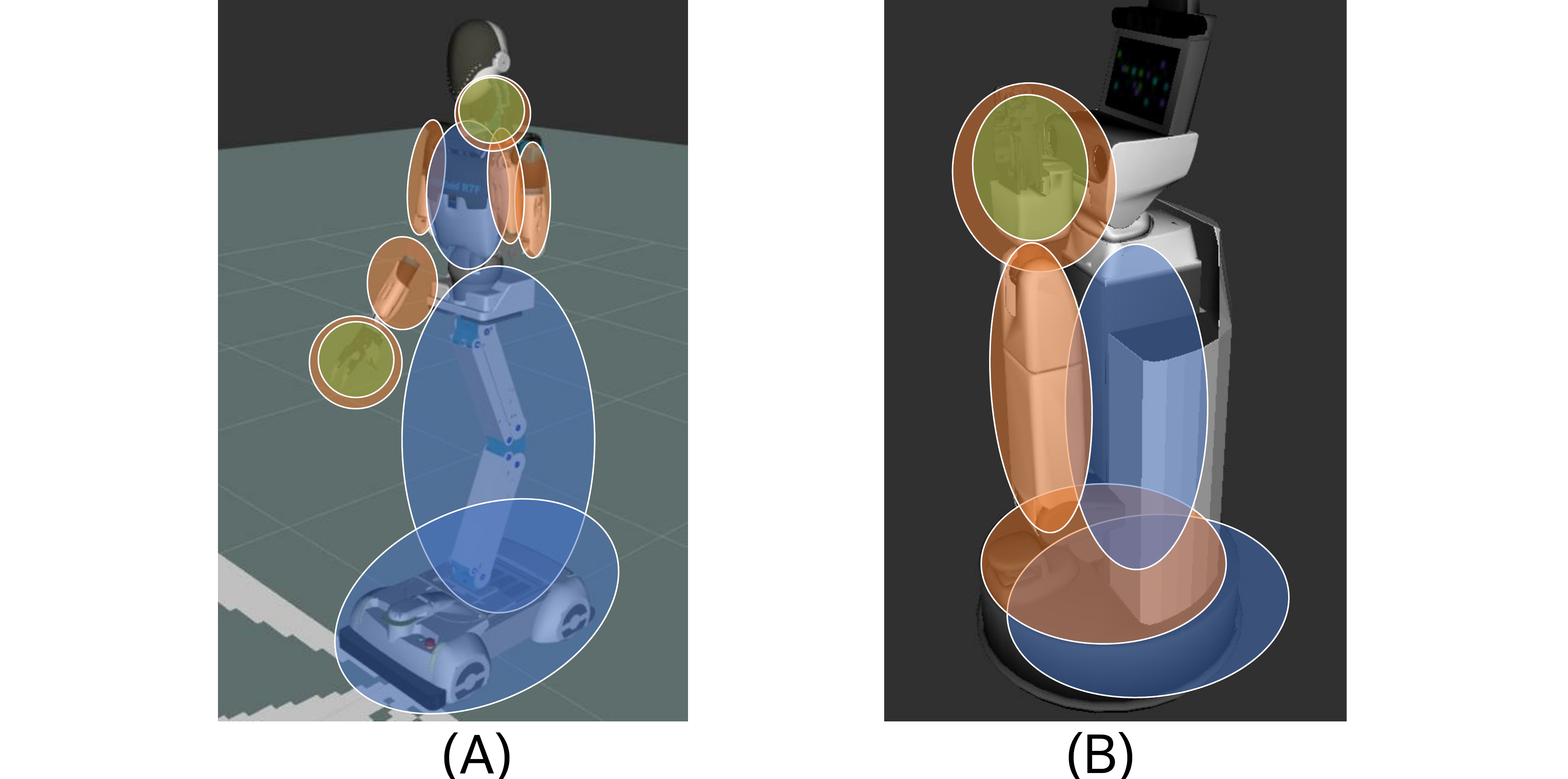}
      \caption{Role division for different robots (A) equal DoF (SEED-noid) (B) fewer DoF (HSR).
Configurational group in orange, positional group in blue, and orientational group in green.}
      \label{various_robots}
   \end{figure}

The mapping design of a named human arm posture to a mapped configuration $\bm{{q_0}^c}$ depends on the number of links (excluding the end-effector) that compose the robot arm. We define two patterns (\figref{various_robots}): the \textit{equal degrees of freedom (DoF)} case where there are exactly two links (same as the human demonstrator) and the \textit{fewer DoF} case where there is only one link.
We will assume that the length of
each robot arm link is nearly equal to the human upper arm or forearm.

\subsubsection{Equal Degrees of Freedom}
\label{mapping_equal}

Since the number of links is equivalent, a na\"{i}ve mapping approach is to copy the named pointing direction of the human upper arm and forearm, to the upper and lower arm link of the robot. However, this way of mapping has no information on the joint-level interpolation between two mapped configurations, and thus it may lack human motion characteristics. For example, let us say an arm is reaching straight from a bent elbow position. The straight arm is a singular point and depending on the twist amount of the upper arm, different end-effector movements will be generated during the interpolation.

To achieve a smooth interpolating motion or a most-likely collision avoiding motion, we must consider the characteristics of the human arm motion. According to Tadokoro et al. \cite{tadokoro1990analysis}, the upper arm usually does not twist during a straight reaching motion, but rather twists when moving the arm to different heights. Therefore, a mapped configuration must be created in a way such that, 1) the pointing direction is kept as much as possible, but 2) the upper arm does not twist between reaching transitions and only twist when there is transition in the height direction. Since we will be using a finite set of arm postures (as explained in \secref{prerequisites_configuration}), the number of transition patterns is also finite. When a robot cannot precisely copy the pointing direction for one of its arm links (e.g., due to joint limitations), we will prioritize the twist constraint when designing the mapped configuration.

\subsubsection{Fewer Degrees of Freedom}
\label{mapping_less}

One approach to map arm postures to robots that have only one arm link is to sum the named pointing direction of the human upper arm and forearm into one direction \cite{ikeuchi2018describing}. However, while this approach may be suitable for gesture motion, manipulation motion have a slightly different characteristic. That is, collision between the forearm and the environment is avoided by the positioning of the elbow and wrist; thus, a summed pointing direction may miss the collision-avoiding essence. To achieve a mapped configuration that is most likely not under collision, we will mainly map the forearm pointing direction to the arm link and refer the root of the link as the elbow. The assumption that lies here is that the upper arm is mainly used to adjust the forward/outward positioning of the elbow; therefore, such motion essence can be (in most cases) alternatively managed with the positional group as long as the root of the arm link is capable of being elevated or planar positioned. 
Likewise, the mapping scheme is applicable for manipulating an articulated object as this is also the case where the upper arm movement is mostly adjusting the elbow position.
To our survey, only 12 out of the 79 named arm postures lie in an exceptional case requiring an upper arm reach, such as reaching over a table.

To achieve the forearm direction, the arm link must be actuated using a horizontally rotating joint and a vertically rotating joint. For some robots, the horizontal rotation may depend on the rotation of the base, therefore, in addition to the arm link, the virtual base rotation may also be included in the configurational group.

\subsection{Solving the Orientation Goal}
\label{solve_go}

   \begin{figure}[b]
      \centering
      \includegraphics[width=\columnwidth]{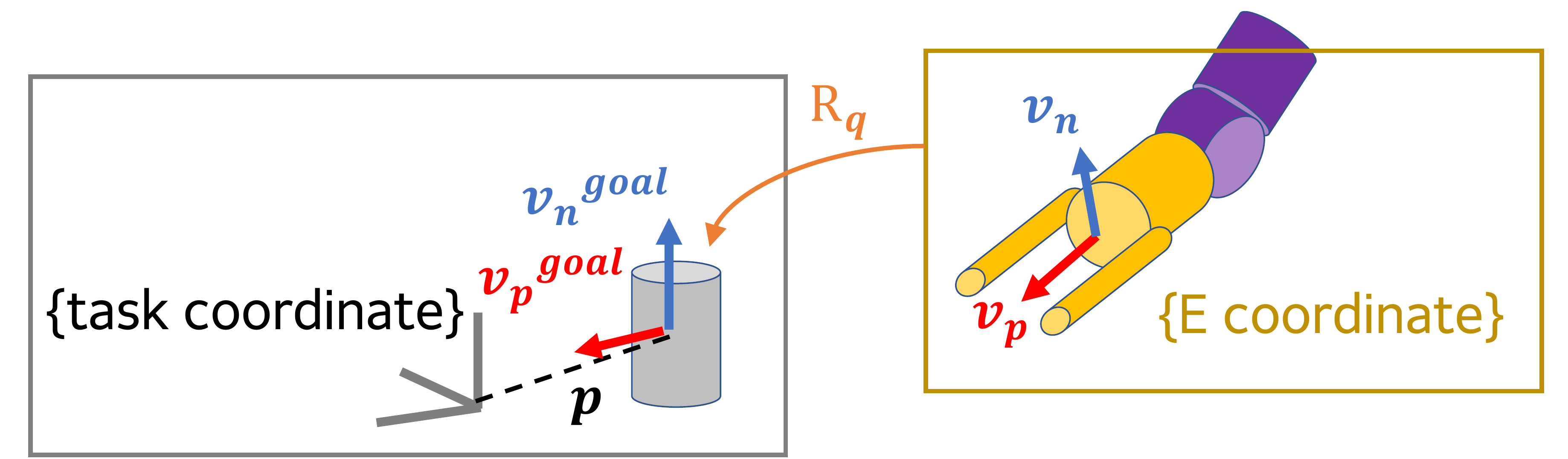}
      \caption{Figure explaining the orientation goal, which uses the palm direction representation taken from the human motion analogy.}
      \label{coordinates}
   \end{figure}

To represent an orientation goal in relation to human mimicking, we use the pointing direction of the palm \cite{guest2014labanotation}. Using this palm analogy, we define a fixed palm unit vector $\bm{v_p}$ on the robot's end-effector $E$ represented in the $E$ coordinate (\figref{coordinates}). The orientation goal is then to point this palm vector toward a desired direction $\bm{{v_p}^{goal}}$ in some fixed task coordinate. With only this condition, the end-effector may take any rotated pose around the palm vector. Therefore, we may choose one fixed perpendicular unit vector $\bm{v_n}$ represented in the $E$ coordinate (usually one that is perpendicular to the line connected by two opposing fingers of a gripper or robotic hand) and make sure $\bm{v_n}$ points to some desired direction $\bm{{v_n}^{goal}}$ in the task coordinate.

An example of $\bm{{v_p}^{goal}}$ is a demonstrated direction such as grasping a can drink from the side or top. An example of $\bm{{v_n}^{goal}}$ is a constrained direction parallel to the axis of a cylindrical handle.

Let  $R_{\bm{q}}$ be a coordinate transformation matrix that transforms $\bm{v_p}$, $\bm{v_n}$ in the $E$ coordinate to the task coordinate when the robot's configuration is $\bm{q}$. Then, using a threshold $\theta_p$, $\theta_n$ the orientation goal is written as below:
\begin{equation}
\label{orinetation_goal}
\Omega_{ogoal}(\bm{q}): \begin{cases}
    1 - \bm{{v_p}^{goal}} \cdot R_{\bm{q}}\bm{v_p}  < \theta_p \\
    1 - \bm{{v_n}^{goal}} \cdot R_{\bm{q}}\bm{v_n}  < \theta_n
  \end{cases}
\end{equation}

\subsection{Solving the Task Position Goal}

Let $\bm{p}$ be the desired position of the end-effector in the task coordinate, and $h(\bm{q_s})$ the end-effector position when the robot's configuration is a sampled configuration $\bm{q_s}$ (calculated using forward kinematics). Then, using a threshold $d$ the task position goal is written as below:
\begin{equation}
\label{task_position_goal}
\Omega_{pgoal}(\bm{q_s}): \|h(\bm{q_s}) - \bm{p}\| < d
\end{equation}

We apply two constraints while solving this task position goal. One is a configuration constraint which ensures that the joint values of the configurational group is kept near the values of the mapped configuration from step 1 and 2. The other is a group connection constraint: when the links actuated by the positional group are the parent or child of the configurational group, a change in value in the positional group may change the look of the links (pointing directions) actuated by the configurational group. The group connection constraint ensures that such situation is avoided.

\subsubsection{configuration constraint}

Let $\bm{{q_s}^c}=\{{{q_s}^c}_i|i=1,...\}$ be the configurational group of a sampled configuration and ${{q_s}^c}_i$ be the $i$-th joint value. Let $\bm{{q_1}^c}=\{{{q_1}^c}_i|i=1,...\}$ be the configuration solved in step 2, and $d_c$ some defined threshold. Then, the configuration constraint is written as below:
\begin{equation}
\label{configuration_constraint}
\Omega_{ccons}(\bm{{q_s}^c}): \sum_{i} | {{q_s}^c}_i - {{q_1}^c}_i  | < d_c
\end{equation}

\subsubsection{group connection constraint}

One way to solve the group connection constraint
 is to use a similar strategy as $\Omega_{ccons}$. Let $L$ be a subset of the positional group that influences the look of the links actuated by the configurational group. Let $\bm{{q_0}^L} \in \bm{{q_0}^p}$ be a partial configuration of the joint configuration from step 1. The subset positional group in a sampled configuration $\bm{{q_s}^L}=\{{{q_s}^L}_i|i=1,...\}$ is kept close to $\bm{{q_0}^L}=\{{{q_0}^L}_i|i=1,...\}$ within a threshold $d_p$ using below:
\begin{equation}
\label{position_joint_constraint}
\Omega_{pcons}(\bm{{q_s}^L}): \sum_{i} | {{q_s}^L}_i - {{q_0}^L}_i  | < d_p
\end{equation}

\section{Using Recordings}
\label{application}

\begin{figure}[t]
      \centering
      \includegraphics[width=\columnwidth]{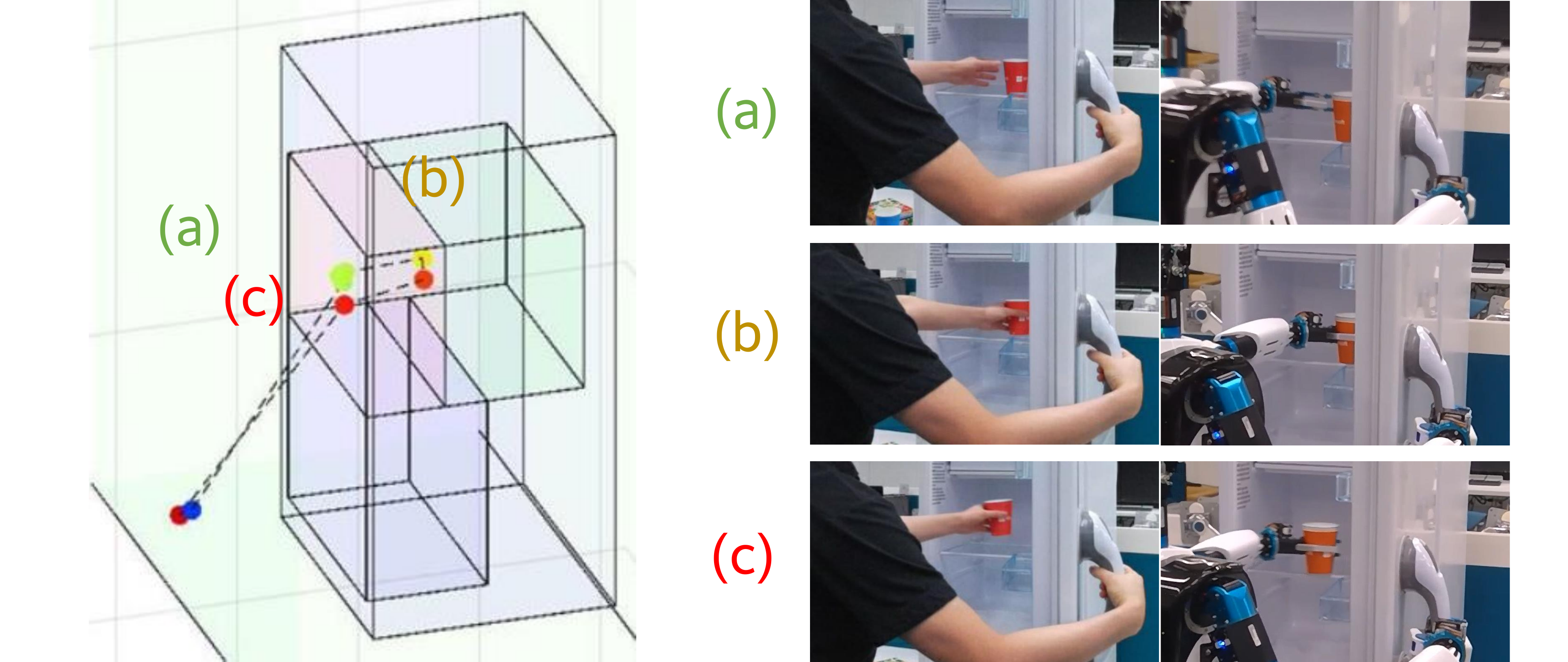}
	 \caption{Left, the detected human hand positions at key visiting waypoints (connected with a dotted line to express time relations) during a reaching and picking task. The image shows three visiting points extracted from the recognized/modelled task environment: (a) point entering the fridge, (b) point of grasping a verbally instructed object, and (c) point of exiting the fridge. Right images show the corresponding arm motion of the human and the robot at these points.}
      \label{input_nc_nc}
   \end{figure}

\begin{figure}[t]
      \centering
      \includegraphics[width=\columnwidth]{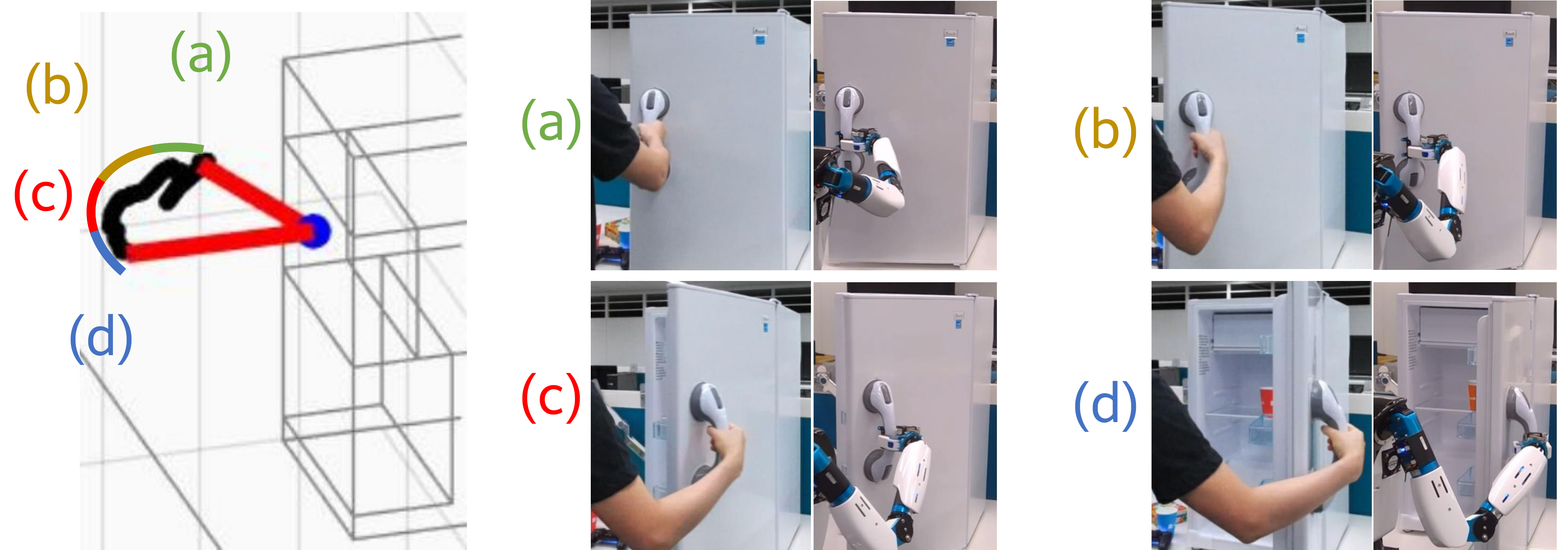}
	 \caption{Left, the detected raw human hand trajectory of a door-opening task (the trajectory after the human grasped the door). Regions (a), (b), (c), (d) indicate where the named arm posture does not change, and the right images show the arm motion of the human and the robot at each of the four regions.}
      \label{input_rv_rv}
   \end{figure}

We extend our discussion of applying our body role division method on a single data point to a series of data obtained in an actual demonstration. The full recording of the demonstrated arm postures (which is then named to represent an arm posture goal) must be aligned with the corresponding task constraints in the instructed task sequence. 
Luckily, our constraints are all related to the actions of the end-effector, therefore, the alignment between the demonstration and constraints is possible by looking at when the human hand trajectory visited the constraint goal values defined for each task (e.g., position of a grasping target, fitting to a door opening trajectory). \figref{input_nc_nc} and \figref{input_rv_rv} show some examples of the aligning in an ``open and pick from a fridge" recording.


One thing to note is that the demonstrated arm posture is suitable only when the robot performs the task under a similar environment to when the demonstration was held. As a limitation of this paper, we will only consider the case where the positional change of an object in an environment is slight (which is the case when picking a mostly-same-place-located item from a fridge). We evaluate in the experiment section on the re-usability of the demonstrated posture under such conditions.

Another thing to note is that the task constraint goal values to visit could be sparse or dense. A sparse example is a reaching or picking task. The robot is able to achieve the task as long as the key waypoints are visited, and collision is avoided. A dense example is a door-opening task. The robot end-effector must follow the exact positions on a specified door-opening trajectory.

In the dense example,
directly using the posture name of a discretized posture as the arm posture goal will have an issue. The posture rarely changes with this naming strategy (\figref{input_rv_rv}) and may generate sudden jumps between postures. To prevent such issues, we represent the arm posture goal in 
the dense case
with the name of the starting posture, ending posture, and an interpolation parameter $t$. To obtain the mapped configuration, the start and end posture is first mapped to configurations $\bm{{q_{(a)}}^c}$, $\bm{{q_{(b)}}^c}$, and an interpolated configuration is calculated as $(1-t)\bm{{q_{(a)}}^c}+t\bm{{q_{(b)}}^c}$.

\section{Experiment}
\label{experiment}

In this section, we evaluate our method on three different task sequences. The first sequence is ``S1: pick from a fridge" consisting of three to four tasks: \textit{``T1: reach for the fridge handle"} followed by \textit{``T2: open the fridge"} and \textit{``T3-4: reach and pick (or look for) a can from inside the fridge."}  The fridge automatically closes if not held, thus one of the robot's arm is always occupied for T3-4. The other sequences include a slightly simpler task of ``S2: locate a target inside a storeroom," which involves operating a prismatic articulated object (i.e., a sliding door), 
and a slightly more difficult task ``S3: take trash when going out." We evaluate S2 
on the HSR robot \cite{yamamoto2019development} (fewer DoF robot), S3 with the SEED-noid robot \cite{sasabuchi2018seednoid} (equal DoF robot), and S1 with both. Since S1 is one of the complex examples that scales for both the equal and fewer DoF robot, we discuss S1 in most detail.

The evaluation was done in a virtual simulator using ROS RVIZ and fake controllers provided by MoveIt. All base movement errors were ignored in this condition. For all the task sequences, the geometric model parameters of the articulated and target object were assumed to be known, and only one demonstration was used to compute the motion.

\subsection{S1: Pick from a Fridge}

\begin{figure}[b]
      \centering
      \includegraphics[width=\columnwidth]{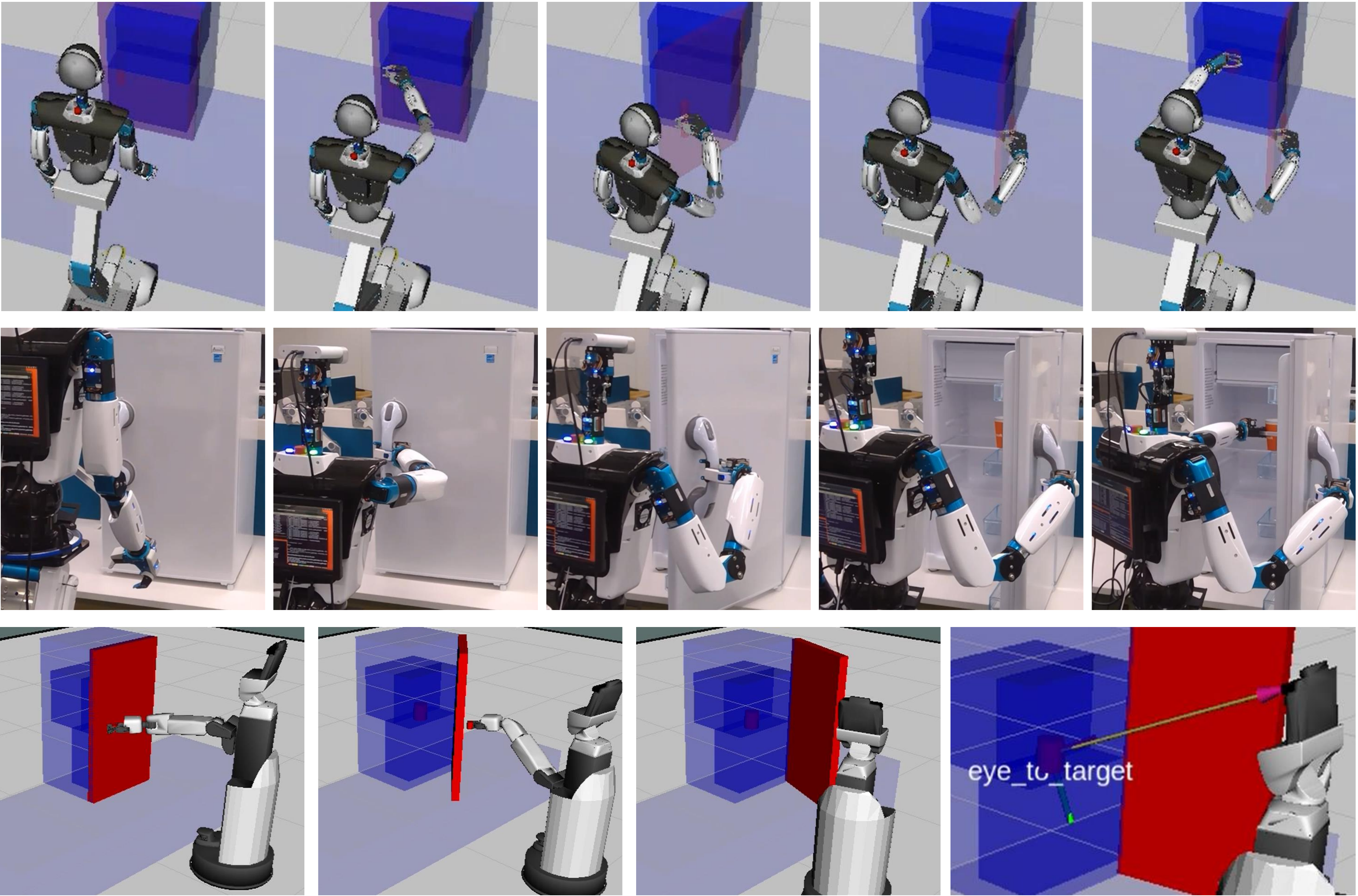}
      \caption{Execution of S1 using our proposed method. Top row: simulated equal DoF. Middle row: real equal DoF. Bottom row: simulated fewer DoF.}
      \label{output_dual}
   \end{figure}

Following the example in \secref{application}, the task position goals for T1 and T3-4 were represented using visiting points. The task position goals for T2 were represented as a trajectory divided into waypoints-to-follow for every 0.1 [radian] opening of the door with a tolerance of 3 [mm] in each direction.
Following the notations in \secref{solve_go}, the orientation goal for T1 and T2 used a direction within 45 degrees of the direction perpendicular to the door plane for $\bm{{v_p}^{goal}}$, and a direction parallel to the door handle axis for $\bm{{v_n}^{goal}}$. For T3-4, a demonstrated approach direction was used for $\bm{{v_p}^{goal}}$, and a direction parallel to the target object's (can's) axis was used for $\bm{{v_n}^{goal}}$. The arm posture goals were obtained 
using an Azure Kinect depth sensor,
and an interpolation parameter as described in \secref{application} was used for T2.
For the configuration constraint threshold, we have used an empirical factor 0.04 (2 degrees tolerance) multiplied by the number of joints in the configurational group. Both robots had 3DoF in the wrist.
\figref{output_dual} shows an execution of the task sequence using our proposed body role method on the simulated fewer and equal DoF robot and the real equal DoF robot.

\subsubsection{Comparison to baseline}

We compared our method with a baseline method which does not use any human demonstration. For the baseline method, we used the default inverse kinematics solver provided by each robot. These solvers search from a current configuration $\bm{q_{t-1}}$ and finds a new configuration $\bm{q_t}$ satisfying the task position and orientation goal. The SEED-noid searches a configuration by solving the position and orientation fitness function (Equation 1, 2) using bio-IK \cite{starke2017memetic} and uses the base when failing to find a solution with the real joints (joints excluding virtual base joints). The HSR searches a configuration by setting the elbow and base yaw as a variable and analytically solves the values of the other joints by prioritizing the base movement \cite{yamamoto2019development}. The main differences between the baseline method and our proposed method are: 1) instead of using the current configuration $\bm{q_{t-1}}$ the solver searches from a configuration $\bm{{q_t}^c}$ mapped from a single human demonstration, 2) to mimic the human demonstration, the solver solves the task position goal and orientation goal as separate steps as explained in \secref{method}, and 3) due to the nature of our method, the choice of the base movement relies on the inverse-reachability of the mapped arm configuration. In addition to the baseline method, we also compared our method with two other methods: A \textit{baseline'} method, which searches from $\bm{{q_t}^c}$ instead of $\bm{q_{t-1}}$ (removes difference 1 between the baseline and our method), and a \textit{retarget} method, which sets the new configuration to $\bm{{q_t}^c}$ without any search (the retargeted motion from a single demonstration).

For this experiment, we used one representative demonstration (same among the two robots). The starting conditions were identical for the compared methods. The starting real joint configurations of the robot presented a mapping of an arm placed downward. The starting base joint configuration (base position) were all set to one that was calculated using our proposed method when executing T1 (we used the one from our method as the baseline solvers required explicitly defining an initial base position and could not calculate one on its own).

We evaluated the motion continuity, environmental collisions, and task achievement of each method. A motion was marked as non-continuous if the robot gripper departed from the door handle position during an interpolation of two solved configurations. A motion was marked as collided if a collision was detected between the fridge (including the door) and any part of the robot's body. A task was marked as achieved if the robot succeeded in picking the can for the equal DoF, and if succeeded in looking at the can without occlusion for the fewer DoF.

\begin{table}[b]
\caption{Comparison of methods in S1 on one representative trial.}
\vspace{-3mm}
\label{result_comparison}
\begin{center}
\begin{tabular}{lcccc}
\hline
equal DoF & baseline & retarget & baseline' & proposed \\
\hline
motion continuity & no & yes & no & yes \\
collisions & no & yes & no & no \\
task achievement & failed & failed & failed & succeeded \\
\hline
fewer DoF & baseline & retarget & baseline' & proposed \\
\hline
motion continuity & no & yes & yes & yes \\
collisions & no & yes & no & no \\
task achievement & failed & failed & failed & succeeded \\
\hline
\end{tabular}
\end{center}
\end{table}

\tabref{result_comparison} shows the results indicating that our method performs better than the other methods in terms of motion continuity and task achievement for both DoF cases. For the equal DoF, the cause of the different results across methods can be explained for the following
reasons.
By dividing the joints contributing to the arm posture and the joints contributing to the task goal, we are able to define a metric (\eqqref{configuration_constraint}) for deciding whether a configuration deviates from the demonstrated motion. Making sure that there is no deviation (and since motion continuity between likely transitions is guaranteed by the mapping scheme), we are able to avoid jumping configurations and also limit the configurations at the end of T2.
The baseline solvers get stuck at a local minimum
while trying to solve the orientation goal. This leads to jumping configurations but also configurations not acceptable for solving T3-4 (e.g., left arm away from the fridge).
Similarly, the baseline solvers for the fewer DoF fails, as it finds a configuration that is optimal for T2 but is not suitable for T3-4. The baseline solver finds a solution closest to the current configuration. This results in a far and slightly-to-the-side base position and occludes seeing inside the fridge. In contrast, our proposed method allows the robot to look inside the fridge from a close and in-front position (\figref{output_dual}).

\subsubsection{Multiple executions}

\begin{table}[t]
\caption{Performance on three demonstrations in the S1 sequence.}
\vspace{-6mm}
\label{result_demo}
\begin{center}
\begin{tabular}{lcccccc}
\hline
  &  \multicolumn{3}{c}{equal DoF} & \multicolumn{3}{c}{fewer DoF}\\
\hline
  & d1  & d2  & d3 & d1 & d2  & d3 \\
total achievements  & 36/50 & 40/50 & 39/50 & 50/50 & 50/50  & 50/50 \\
\hline
\end{tabular}
\end{center}
\end{table}

   \begin{figure}[t]
      \centering
      \includegraphics[width=\columnwidth]{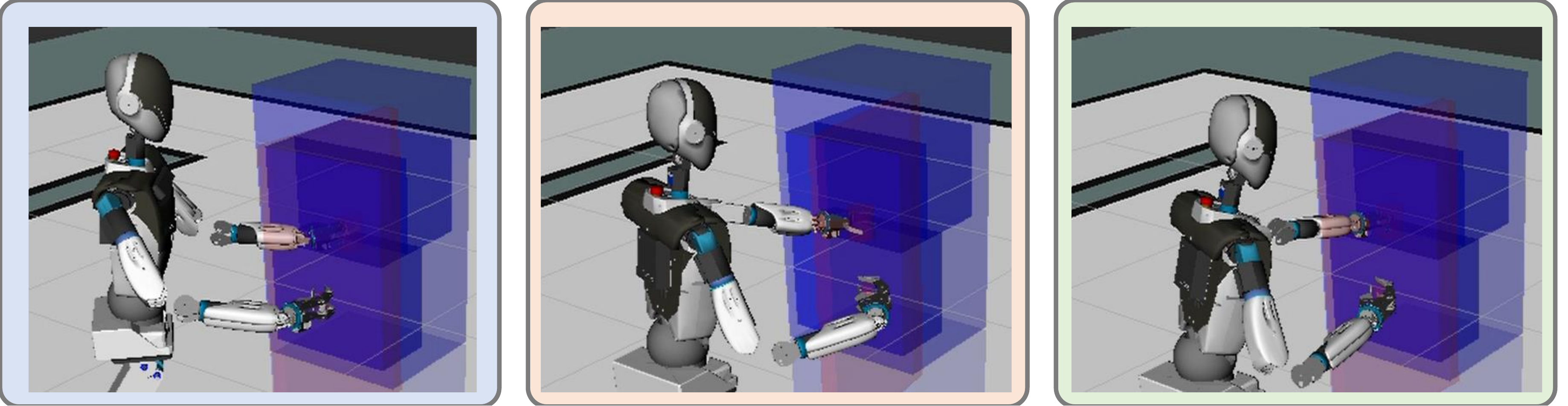}
      \caption{Execution of demonstrations from three different starting configurations by the equal DoF robot. Left d1, middle d2, right d3.}
      \label{demos}
   \end{figure}

To better evaluate our method on the robustness against slight demonstration variances, we tested three different demonstrations (d1, d2, d3)
where the human demonstrated from a different standing position for d1, and similar positions but a different starting configuration for d2 and d3
(\figref{demos} shows an execution of the three demonstrations by the equal DoF robot).
For each demonstration, we performed 5 executions of T2 in simulation and obtained five slightly different T2-ending configurations. Then, on each of the five obtained configurations, we tested the performance of T3-4 on 10 varying target object locations (five locations inside the fridge sampled left to right for a near and slightly far position) and counted the number of successful achievements. This totals to fifty executions per demonstration.
The same task achievement conditions from the previous experiment were used for each robot.

\tabref{result_demo} shows that similar results are obtained for all demonstrations. The discretized representation of the arm postures enables the arm motion in T2 and T3-4 to be similar among the different demonstrations. 
The task of the fewer DoF robot was simple (look at the target without occlusion), therefore the robot succeeded in all trials.
However, some executions failed for the equal DoF.
In these executions, the demonstrated reaching motion in T3-4 was not suitable for the target locations (e.g., a reaching motion in a forward direction was required for the location but the demonstration was shown for a different location where the human reached diagonally). The failure cases were mostly when the target location differed more than 15 cm from the demonstration. Yet, when the target object was positioned close to the position as demonstrated (which covered more than half of the fridge's width), our method was capable of adjusting the mapped configuration to achieve the grasping under slight positional changes.

   \begin{figure}[t]
      \centering
      \includegraphics[width=\columnwidth]{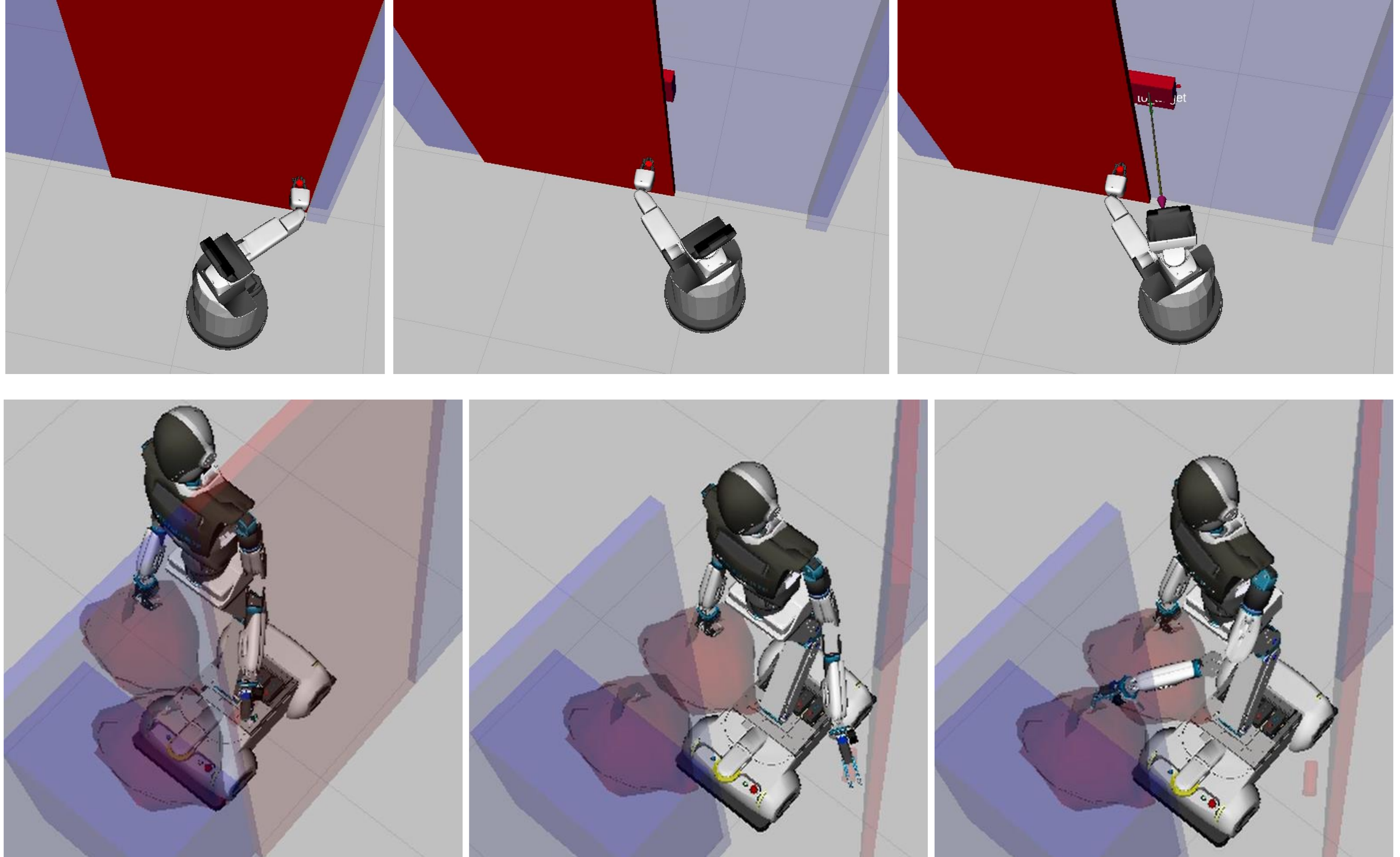}
      \caption{Execution of S2 (top row) and S3 (bottom row) using our proposed method.}
      \label{s2_s3}
   \end{figure}

   \begin{figure}[t]
      \centering
      \includegraphics[width=\columnwidth]{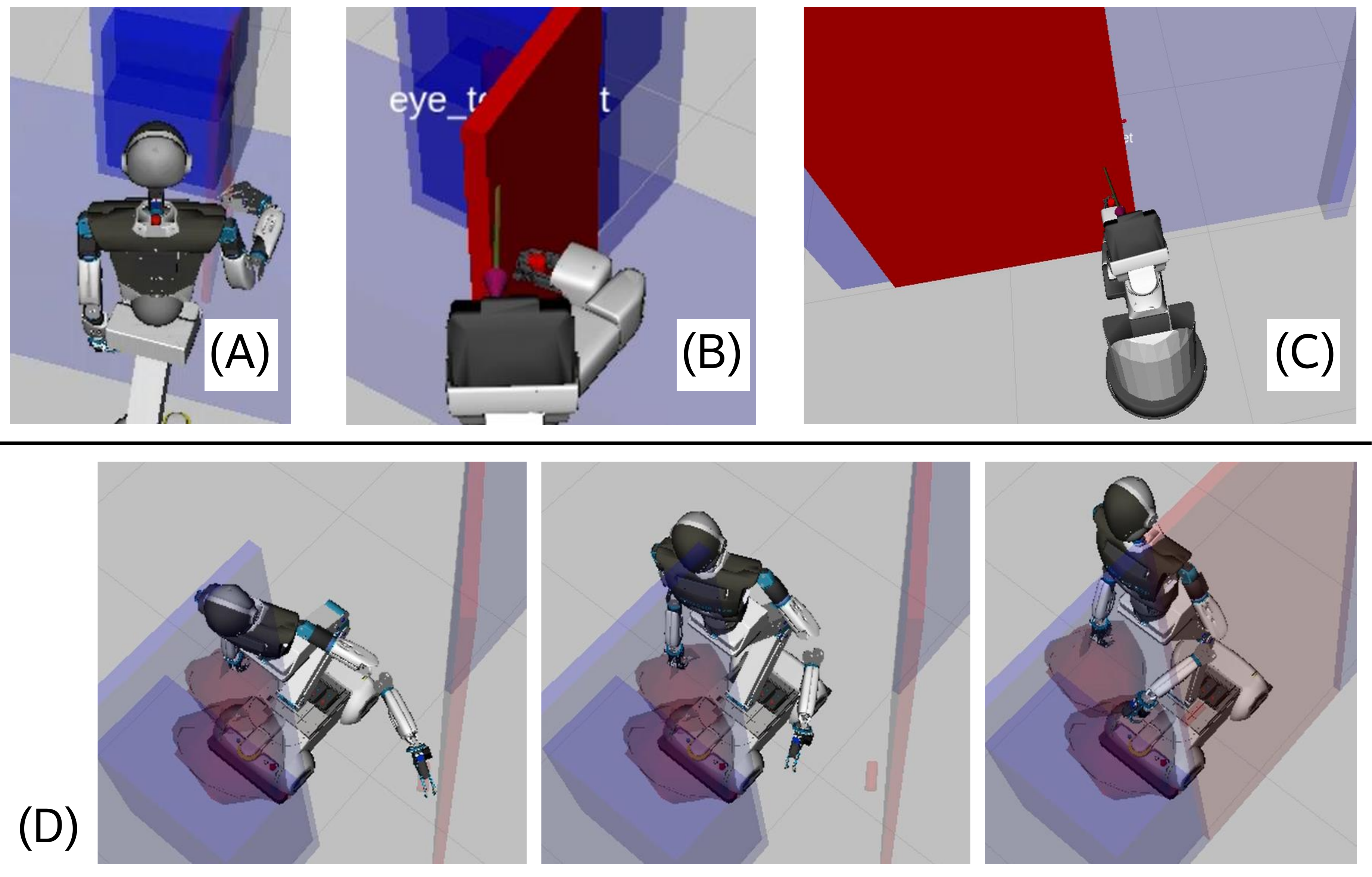}
      \caption{Failure cases of the baseline method. (A) S1 equal DoF shifted to the right of the fridge. (B) S1 fewer DoF fails to look inside due to occlusion. (C) S2 fewer DoF fails to locate the target as occluded. (D) S3 equal DoF opens the door but the door closes when the gripper is released as nothing is supporting the door.}
      \label{failures}
   \end{figure}

\subsection{S2 and S3}

An execution by the proposed method for sequences S2 and S3 are shown in \figref{s2_s3} and failures by the baseline are shown in \figref{failures}. Similar to S1, the baseline planner in S2 finds an optimal solution for manipulating the sliding door but fails to achieve the following task of locating the target inside.
With the baseline, the robot fails to follow the intended human instructions of \textit{open and look}.
Meanwhile, S3 is composed of the same tasks as S1 \textit{reach}, \textit{open}, and \textit{pick} but with completely different constraints when they are sequenced. The robot's right arm is occupied and therefore must both open and grasp the target with its left arm. This forces the robot to release the door at grasping the target. To accomplish this task, the robot must first open the door in a way that the robot's body is kept close to the door. Then, hold the door opened with its body (i.e., base). Due to the inverse reachability of the mapped configuration, our method is able to implicitly achieve such a state and also achieve the following reaching and picking task. However, the baseline fails as it finds a local optimum solution which opens the door by only extending the robot's arm far away. With this state, the door closes once the robot releases the door.

\section{Conclusion}
\label{conclusions}

Applying both
task constraints and implicit knowledge from motion
is essential for solving a complex task sequence. The task constraints lack information on how to execute a task with the entire task sequence in mind unless modeled by an expert. The motion knowledge lacks information on how to scale the motion in a way that achieves the task objective unless multiple demonstrations are provided. The two knowledge are applied by dividing a robot's body configuration into configurational, positional, and orientational groups. Only one human demonstration is needed as long as the structure of the robot is capable of mapping the arm motion analogy.

Our method is potentially beneficial for a range of tasks where the arm motion provides hints on the task-to-task constraints.
These constraints differ even for the same combination of tasks and are dependent on the context or intent of the sequence.
Despite the different intents, our method was capable of foreseeing the reachability/eyesight and positioning required for an upcoming task in each sequence by limiting/guiding the configurations to ones that were appropriate. Such results were achieved for both the equal and fewer DoF robot and the results were more stable compared to the pure IK solvers.
A few limitations of our current work are that the tasks are very close to the original demonstrations, the demonstration did not involve a large movement in the trunk (such as bending toward the floor), and the demonstrations did not involve footsteps that rotated the entire body.
Also, achieving every intent of the task sequence is not always guaranteed, which is a common issue in motion-based approaches. These are future directions for extending our work.


%





\ifCLASSOPTIONcaptionsoff
  \newpage
\fi



%



\bibliographystyle{IEEEtran}
\bibliography{bib}

%








\end{document}